# Particle Swarm Optimization Based Reactive Power Optimization

P.R.Sujin, Dr.T.Ruban Deva Prakash and M.Mary Linda

**Abstract**— Reactive power plays an important role in supporting the real power transfer by maintaining voltage stability and system reliability. It is a critical element for a transmission operator to ensure the reliability of an electric system while minimizing the cost associated with it. The traditional objectives of reactive power dispatch are focused on the technical side of reactive support such as minimization of transmission losses. Reactive power cost compensation to a generator is based on the incurred cost of its reactive power contribution less the cost of its obligation to support the active power delivery. In this paper an efficient Particle Swarm Optimization (PSO) based reactive power optimization approach is presented. The optimal reactive power dispatch problem is a nonlinear optimization problem with several constraints. The objective of the proposed PSO is to minimize the total support cost from generators and reactive compensators. It is achieved by maintaining the whole system power loss as minimum thereby reducing cost allocation. The purpose of reactive power dispatch is to determine the proper amount and location of reactive support. Reactive Optimal Power Flow (ROPF) formulation is developed as an analysis tool and the validity of proposed method is examined using an IEEE-14 bus system.

——————————— ◆ ———————————

## 1 INTRODUCTION

Real time pricing of electrical energy is an area of intense research at present. Real time pricing of reactive power is closely related to that of active or real power.

Reactive power plays a significant role in supporting the real power transfer, which becomes especially important ,thereby increasing number of transactions are utilizing the transmission system and voltages becomes a bottleneck in preventing additional power transfer.

Many real time pricing methods were established. Lamont JW and FU J[l] proposed the importance of reactive power voltage support. Dai Y et al [2] proposed a sequential Quadratic programming method for reactive power pricing. Bhattacharya K and Zhong J [3] proposed the problem of reactive power procurement by an Independent System Operator (ISO) in deregulated electricity markets.

Baughman ML and Siddiqi SN [4] presented an analysis made of real time pricing policies of reactive power using a modified OPF model. Li YZ and David AK. [5] extended wheeling rates of real power to include reactive power transportation using appropriate AC OPF model.

Caramanis MC et al [6] developed a theory for spot price of electricity. Ei-keib AA and Ma X [7] proposed the proper pricing of active and reactive power for economic mid secure operations of power systems in an open transmission access environment.

Baughman ML et al [8] have proposed a mathematical formulation model for real time pricing of electricity that included selected ancillary services. Further developed a competitive market mechanism based on it in [9]. However, as pointed out in [10] the application of marginal reactive price is not very practical due to its volatile and erratic behaviors.

Hao S [11] suggested that the management of reactive resources in particular the generation facilities under control of transmission operators, plays an important role in maintaining voltage stability and system reliability.

Silva EL et al [12] addresses both the principles and practical issues involved in developing cost based payments for reactive power.

Li YZ and David AK [13] proposed wheeling in the transmission of electrical power and reactive power by a seller to a buyer through a transmission network owned by a third party.

Singh C and Musavi MT [14] proposed an "energy function" for transient stability analysis of power systems.

James Kennedy, Russell Eberhart [15] introduced Particle Swarm Optimization algorithms. Zwe-Lee Gaing [16] proposed Particle Swarm Optimization based unit commitment algorithm.

The objective of this paper is to minimize the cost of to-

————————————————

- P.R.Sujin[1,] Research Scholar, Dept.of Electrical and Electronics Engineering, NI University, Kumaracoil, Kanyakumari District, Tamil Nadu,India.
- Dr.T.Ruban Deva Prakash[2], Prof. Dept.of Electrical and Electronics Engineering, NI University, Kumaracoil, Kanyakumari District, Tamil Nadu, India.
- M.Mary Linda[3] Asst. Professor ,Dept. of Electrical and Electronics Engineering,Ponjesly College of Engineering,Nagercoil, Kanyakumari District,Tamil Nadu, India.

tal reactive support from generators and reactive compensators and find the payment to the same. In this paper Particle Swarm Optimization is used for optimizing the cost of reactive power.

## 2 PROBLEM FORMULATION

### 2.1 Reactive Cost of Generators

A generator's capacity constraint, which is usually called the loading capability, plays an important role in calculating its opportunity cost. Generators provide reactive support by producing or consuming reactive power, which can be represented by operating with lagging or leading power factors.

Opportunity cost also depends on the real time balance between demand and supply in the market. The model for opportunity cost is presented as follows.

$$C_{gqi}(Q_{gi}) = [C_{gqi}(S_{gi\,max}) - C(\sqrt{S^2_{gi\,max} - Q^2_{gi}}\, K_{gi})] \quad (1)$$

where,

$Q_{gi}$ - the reactive power output of generator $g_i$,
$S_{gi\,max}$ - the maximum apparent power of generator g
$C_{gpi}$ - the active power cost which is modeled as a Quadratic function.
$P_{gi}$ - the active power output of $g_i$ a, b, and c are cost coefficients
$k_{gi}$ - an assumed profit rate for active power generation at bus i.

### 2.2 Cost of Reactive Compensators

A compensator may be considered to be source of reactive power reserve, whose main function is non-control voltage profile during transient periods. The charge for using reactive compensators is assumed proportional to the amount of the reactive power purchased and can be expressed as:

$$C_{cj}(O_{cj}) = r_j Q_{cj} \quad (2)$$

where,

$r_j$ - the reactive cost,

$Q_{cj}$ – the reactive power purchased.

The production cost of a compensator is assumed as its capital investment return, which can be expressed as its depreciation rate. For example, if the investment cost of a reactive compensator is $6200/MVAr, and its average working rate and life span are 2/3 and 30 years respectively, the cost or depreciation rate of the compensator can be calculated as:
$r_j$ = investment cost/operating hours
   = ($6200)(30x365x24x(2/3)=$0.0354/MVARh      [17]

## 3 REACTIVE ANCILLARY SERVICE PROCUREMENT

With a reactive bid structure established, the Independent System Operator [ISO] requires a proper criterion to determine the best offers and hence formulate its reactive power procurement. The purpose of reactive power dispatch is to determine the proper amount and location of reactive support in order to maintain a proper voltage profile and voltage stability requirement.

### 3.1 Reactive Optimization model

*Objective:*

The suggested objective function is:

$$\text{Min } C_Q = \sum_{i=NG} C_{gqi}(Q_{gi}) + \sum_{i=NC} C_{ci}(Q_{ci}) \quad (3)$$

where,

$C_Q$ - the total reactive support cost from generators and reactive compensators

NG - the set of all generator buses

NC - the set of all reactive compensator buses.

**Constraints in OFF**

The equality constraints of OFF problem are load flow equations.

$$P_{gi} = \left|\overline{V_i}\right| \sum_{j=N} \left|\overline{V_j}\right| \left|\overline{Y_{ij}}\right| \cos(\theta_{ij} + \delta_j - \delta_i) \quad (4)$$

$$P_{li} = \left|\overline{V_j}\right| \sum_{j=N} \left|\overline{V_i}\right| \left|\overline{Y_{ij}}\right| \cos(\theta_{ij} + \delta_j - \delta_i) \quad (5)$$

$$\theta_{gi} = \left|\overline{V_i}\right| \sum_{j=N} \left|\overline{V_j}\right| \left|\overline{Y_{ij}}\right| \sin(\theta_{ij} + \delta_j - \delta_i) \quad (6)$$

$$\theta_{li} = \left|\overline{V_j}\right| \sum_{j=N} \left|\overline{V_i}\right| \left|\overline{Y_{ij}}\right| \sin(\theta_{ij} + \delta_j - \delta_i) \quad (7)$$

where,

N - total number of buses in the system

$P_{Li}$ & $Q_{Li}$ - the specified active and reactive demand at load bus i

$Y_{ij} \angle \theta_{ij}$ - the element of the admittance matrix

$V_i = V_i \angle \delta_i$ - the bus voltage at bus i

The inequality constraints are:

$$V_{i,min} \leq \left|\overline{V_i}\right| \leq V_{i,\,max} \quad (8)$$

$$Q_{gi,min} \leq Q_{gi} \leq Q_{gi,max} \quad (9)$$

$$Q_{ci,min} \leq Q_{ci} \leq Q_{ci,max} \quad (10)$$

$V_{i,min}$ and $V_{i,\,max}$ - the lower and upper limits of bus voltage

$Q_{gi,min}$ and $Q_{gi,\,max}$ - the lower and upper limits of reactive power output of the generator

$Q_{Ci,mjn}$ and $Q_{Ci,max}$ - the lower and upper limits of reactive

power output of the compensators.[17]

## 4 PARTICLE SWARM OPTIMIZATION (PSO) ALGORITHM PROCEDURE

The PSO iteration is carried out to obtain the reactive power minimization as shown in the flow chart.

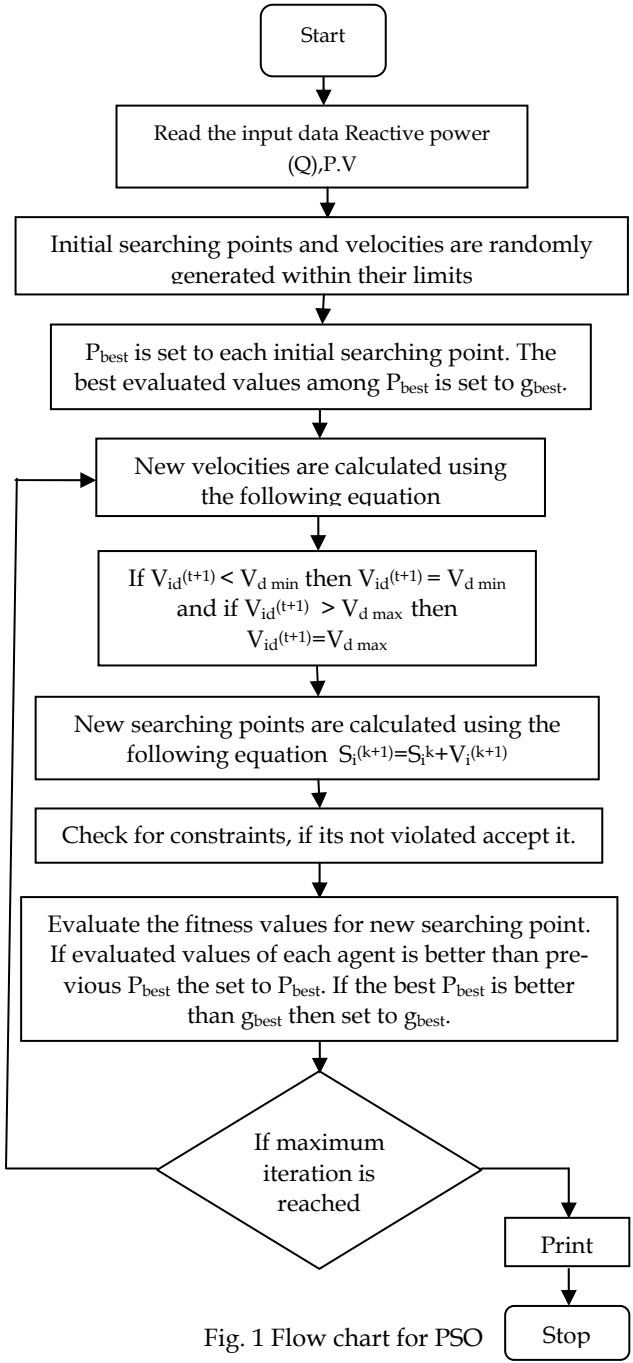

Fig. 1 Flow chart for PSO

1. Initial searching points and velocities are randomly generated within their limits.
2. $P_{best}$ is set to each initial searching point. The best-evaluated values among $P_{best}$ is set to $g_{best}$.
3. New velocities are calculated using the equation
   $V_i^{(k+1)} = W_i \cdot V_i^k + C1 \ast rand() \ast (Pbest_{id} - S_i^k) + C2 \ast rand() \ast (gbest_{id} - S_i^k)$.
4. If $V_{id}^{(t+1)} < V_{d\,min}$ then $V_{id}^{(t+1)} = V_{d\,min}$ and if $V_{id}^{(t+1)} > V_{d\,max}$ then $V_{id}^{(t+1)} = V_{d\,max}$.
5. New searching points are calculated using the equation $S_i^{(k+1)} = S_i^k + V_i^{(k+1)}$.
6. Evaluate the fitness values for new searching point. If evaluated values of each agent is better than previous $P_{best}$ then set to $P_{best}$. If the best $P_{best}$ is better than best $g_{best}$ then set to $g_{best}$.
7. If the maximum iteration is reached stop the process otherwise go to step3.

## 5 IMPLEMENTATION OF PSO

The Particle Swarm Optimization technique is implemented using Matlab7 and is tested on an IEEE 14 bus system. The optimization problem considered in this case is to minimize the total reactive support cost from generators and reactive compensators. The reactive power injection is used as encoding particles. The objective function in this optimization problem is used as a fitness function in the PSO. The following combination of control parameters are used for running the PSO. The inertia weight in the range 0.9 to 1.2 on an average has a better performance, and has a large chance to find the global optimum within a reasonable number of iterations. Using the above parameters the PSO is executed and the results are obtained.

### 5.1 Development of Algorithm

Step1. Perform the optimal power flow

Step2. Reactive power is taken as the initial population

Step3. Choose the population size and number of gener-Ation

Step4. Select the reactive power injection as state variable ($X_i$)

Step5. Initial searching points and velocities are randomly Generated with in their limits

Step6. $P_{best}$ is set to each initial searching point. The best evaluated values among $P_{best}$ is set to $g_{best}$

Step7. New velocities are calculated using the equation
$V_{id}^{(t+1)} = W_i \cdot V_{id}^t + C1 \ast rand() \ast (Pbest_{id} - X_{id}^{(t)}) + C2 \ast rand() \ast (gbest_{id} - X_{id}^{(t)})$

Step8. If $V_{id}^{(t+1)} < V_{d\,min}$ then $V_{id}^{(t+1)} = V_{d\,min}$ and if

$V_{id}^{(t+1)} > V_{d\ max}$ then $V_{id}^{(t+1)} = V_{d\ max}$

Step 9. New searching points are calculated using the Equation $S_i^{(k+1)} = S_i^k + V_i^{(k+1)}$

Step 10. Evaluate the fitness values for new searching point according to the objective function given below

$$\text{Min } C_Q = \sum_n^{i=j} C_{gqi}(\theta_{gi}) + \sum_n^{l=j} C_{ci}(\theta_{ci})$$

If evaluated values of each agent is better than

Previous $P_{best}$ then set to $P_{best}$. If the best $g_{best}$ is better than best $P_{best}$ then set to $g_{best}$.

Step 11. Stop criteria. Maximum number of generation is reached or optimal point is achieved.

Step 12. To computes total power loss before compensation and after compensation.

Step 13. To compute total reactive support cost from generators and reactive compensators

Step.14. To find the payment to generators and reactive Compensators

## 6 REACTIVE PRICING SCHEME

In many deregulated markets, the ISO has a limited access to information on generators and hence may not be able to determine a generator's revenue loss. An appropriate option in such markets is to call for reactive bids from generators.

We discussed regions on the reactive power coordinate in the previous section, which are now explicitly defined here to formulate the generator's expectation of payment function. From knowledge of generators expectation of payment, the ISO can call for reactive bids from all parties. With a reactive bid structure established, the ISO requires a proper criterion to determine the best offers and hence formulate its reactive power procurement plan. The ISO determines the marginal benefit of each reactive bid with regard to system losses. The ISO shall seek to minimize losses least; it would require procuring higher loss compensation services (also involving payments).

A novel pricing scheme for reactive power is presented in the section.

1. Re active power support cost responsibility separation. The total reactive power cost is divided into two components, namely the generators side and the loads side. The duty cost of the generators side $C_G$ (i.e. the reactive cost to support the delivery of active power) is circulated as the optimal value of Eq. (3) when the system has no reactive loads. To evaluate this cost, the power factors of all the loads are set to unity. This component of cost is caused only by real power transportation. The remaining cost (CL - CQ* - CG) is assigned to reactive loads.
2. Equitable allocation of $C_G$ to generators.
3. Payment to generators.
4. Payment to independent reactive sources.

TABLE 1
GENERATORS DATA

| Generator Number | $G_1$ | $G_2$ |
|---|---|---|
| Maximum apparent power (p.u) | 0.9 | 0.9 |
| Active power output (p.u) | 0.74 | 0.6 |
| Reactive power limit (p.u) | [-0.5,0.4] | [0.4,0.5] |
| Profit rate (p.u) | 0.07 | 0.07 |
| Active power cost function ($<h) | $45 + 750P_i + 450P_i^2$ | |

TABLE 2
DEPRECIATION RATES OF COMPENSATORS

| Bus number | 3 | 4 |
|---|---|---|
| Maximum capacity (p.u) | 0.3 | 0.3 |
| Depreciation coefficients (S/MVArh) | 0.10 | 0.10 |

TABLE 3
TRANSFORMER DATA

| From | To | Resistance (p.u) | Reactance (p.u) | Ratio (p.u) |
|---|---|---|---|---|
| 4 | 5 | 0.0000 | 0.25202 | 0.932 |
| 7 | 6 | 0.0000 | 0.20912 | 0.978 |
| 9 | 6 | 0.0000 | 0.55618 | 0.969 |

TABLE 4
TRANSMISSION LINE DATA

| From | To | Resistance (P.u) | Reactance (P.u) | Susceptance (p.u) |
|---|---|---|---|---|
| 3 | 2 | 0.04699 | 0.19797 | 0.0219 |
| 3 | 6 | 0.06701 | 0.17103 | 0.0173 |
| 2 | 7 | 0.05695 | 0.17388 | 0.0170 |
| 2 | 6 | 0.05811 | 0.17632 | 0.0187 |
| 2 | 1 | 0.01938 | 0.05917 | 0.0264 |
| 5 | 8 | 0.00000 | 0.17615 | 0.000 |
| 4 | 11 | 0.09498 | 0.19890 | 0.000 |
| 4 | 12 | 0.12291 | 0.25581 | 0.000 |
| 4 | 13 | 0.06615 | 0.13027 | 0.000 |
| 7 | 6 | 0.01335 | 0.34802 | 0.0064 |
| 7 | 1 | 0.05403 | 0.22304 | 0.0246 |
| 8 | 9 | 0.00000 | 0.11001 | 0.000 |
| 14 | 9 | 0.12711 | 0.27038 | 0.000 |
| 14 | 12 | 0.17093 | 0.34802 | 0.000 |
| 9 | 10 | 0.03181 | 0.0845 | 0.000 |
| 10 | 11 | 0.08205 | 0.19207 | 0.000 |
| 12 | 13 | 0.22092 | 0.19988 | 0.000 |

TABLE 5
LOAD DATA

| Bus number | Real load (p.u) | Reactive load (p.u) |
|---|---|---|
| 5 | 0.076 | 0.016 |
| 6 | 0.478 | 0.039 |
| 8 | '0.150 | 0.05 |
| 9 | 0.595 | 0.024 |
| 10 | 0.090 | 0.058 |
| 11 | 0.035 | 0.018 |
| 12 | 0.066 | 0.016 |
| 13 | 0.150 | 0.058 |

TABLE 6
REACTIVE POWER OPTIMIZATION USING PSO

$r_j$ = $0.0354/MVAR- h

Requirement of VAR sources in p.u.

0.1200   0.0700   0.1700   0.5600

Power loss before compensation =0.100682p.u

Power loss after compensation =0.0839515p.u.

Cost of reactive contribution in $/h

0   0.2975   0.6016   1.9816

Payment to generators and reactive

Compensators = 2.8807 $/h

## 7 SIMULATION RESULTS AND DISCUSSION

The objective is achieved by maintaining the whole system total power loss as minimum. Hence minimum cost allocation can be achieved. By supplying reactive power the voltage of the system can be maintained within limits.

The objective function is to minimize the cost of total reactive support from generators and reactive compensators and to find the payment to generators and reactive compensators.

An IEEE-14 bus system is used for the test system. A feature of the system is large power transfers from the top area of the bottom area over a long transmission distance. This makes it appropriate to study reactive power and voltage problems and hence it is adopted in the paper. The system has three generators, 14 buses and 20 tie lines. Two independent reactive compensators IC3 and IC4, are located at bus 3 and bus 4, respectively .bus 14 is selected as slack bus and designated to make good transmission loss changes and its reactive power cost is not included in the optimisation procedure. The system base capacity is 100 MVA.

Table 1 provides generators, Gl and G2 data, which are used for reactive power opportunity cost analysis. Capacity and depreciation coefficient of reactive compensators are listed in Table2. Transformers data, Transmission lines data, and loads data are given in Table 3-5, respectively. *By* increase the load continuously the current increases and the voltage level decreases to make the voltage level constant by increasing the excitation, i.e. increasing the reactive power.

Without reactive power we cannot operate the machines properly. VAR equipments at the load centres, rather than at generators. By changing the taps of transformer the voltage level can be improved. Load shedding is the simplest way to improve the voltage level. Simulation results are obtained by the proposed method for the system data presented (table 6).

## 7 CONCLUSION

In this paper reactive OPF is developed to solve the optimal reactive dispatch problem. The total reactive cost is separated into generators' duty and loadings' duty. Cost duty on the generation side is allocated to real power sellers by evaluating their reactive power requirement for real power transportation. The method of evaluation adopted in this paper has a common basis for every market participant and hence it is consistent and equitable. Each generator will be paid according to the difference between its actual incurred cost of contributing of reactive power support and its cost of reactive power requirement for real power selling. The theory and implementation is illustrated through a simple example. The results are obtained using PSO illustrates that the proposed algorithm is simple and practical. This method is compatible with the new competitive market structure and economic efficiency can be achieved. The algorithm is tested on an IEEE-14 bus system for various control parameters of the Particle Swarm Optimization (PSO). The results obtained give an efficient, feasible and optimal solution.